\documentclass{llncs}
\usepackage[T1]{fontenc}
\usepackage{xcolor}
\usepackage{marvosym}
\usepackage{graphicx}
\usepackage{booktabs}
\usepackage[hidelinks]{hyperref}
\usepackage{cleveref}
\begin{document}
\title{OpenQlaw: An Agentic AI Assistant for Analysis of 2D Quantum Materials}

%

\author{
Sankalp Pandey\inst{1}\textsuperscript{(\Letter)}\orcidID{0009-0003-3140-4710} \and
Thanh-Dat Truong\inst{1}\orcidID{0000-0003-4634-8793} \and
Xuan-Bac Nguyen\inst{1}\orcidID{0000-0001-8495-5469} \and
Hoang-Quan Nguyen\inst{1,4}\orcidID{0009-0009-3522-2563} \and
Tim Faltermeier\inst{2,4}\orcidID{0009-0006-5570-9975} \and
Nicholas Borys\inst{2,4}\orcidID{0000-0001-5434-1191} \and
Hugh Churchill\inst{3,4} \and
Khoa Luu\inst{1,4}\orcidID{0000-0003-2104-0901}
}

\authorrunning{S. Pandey et al.}
%

\institute{
Quantum AI Lab, University of Arkansas,
Fayetteville, AR 72701, USA\\
\email{\{sankalpp,tt032,xnguyen,hn016,khoaluu\}@uark.edu}
\and
Department of Physics and Astronomy, University of Utah,
Salt Lake City, UT 84112, USA\\
\email{timfaltermeier@montana.edu, nicholas.borys@utah.edu}
\and
Department of Physics, University of Arkansas,
Fayetteville, AR 72701, USA\\
\email{hchurch@uark.edu}
\and
MonArk NSF Quantum Foundry, USA\\
\texttt{\url{https://uark-cviu.github.io/projects/OpenQlaw}}
}

\maketitle
\begin{abstract}
Physics-aware multimodal large language models (MLLMs) can localize and characterize two-dimensional (2D) material flakes from optical microscopy images, producing rich candidate observations and scientific reasoning. However, downstream workflows also require reliable operations over these outputs, including counting, selection, measurement, and rendering. We present OpenQlaw, a tool-using agentic artificial intelligence (AI) system that connects a general vision-language orchestrator with Physics-Aware Instruction Tuning (QuPAINT), a physics-aware domain expert. Through evaluation on fully reviewed optical microscopy images, we show that summary counts derived from longer free-form candidate enumerations can become inconsistent with the underlying returned observations. Controlled replay experiments further show that structuring the same observations alone does not resolve this behavior, while deterministic delegation returns the exact requested record count in all cohort trials. To address this, we introduce State-Grounded OpenQlaw, which converts expert outputs into image-scoped observation records and performs deterministic scientific operations over explicitly identified records with refusal conditions when required information is unavailable. This preserves the expert's original analysis while making downstream operations exact over the selected record and independently evaluable from candidate-list discrepancies.
\end{abstract}

\section{Introduction}

Atomically thin two-dimensional (2D) quantum materials such as graphene, MoS$_2$, WSe$_2$, and hBN have become important platforms for electronic, photonic, and quantum devices \cite{nguyen2025quantumbrainquantuminspiredneuralnetwork,nguyen2024quantum,nguyen2025diffusion,nguyen2024hierarchical,dendukuri2019defining,nguyen2023quantum,11250125,holliday2025advancedhybridquantumtabu,holliday2025advancedquantumannealingapproach,11250175,holliday2024hybrid}. Their physical properties depend strongly on material thickness, making reliable identification of exfoliated flakes an important step in device fabrication. In practice, researchers inspect optical microscopy images to locate candidate flakes and estimate their layer thickness from optical contrast and interference color before further characterization \cite{ISLAM2025100161,ng2026high,Harris_2025}.

Recent deep-learning methods increasingly automate this perception task through object detection and segmentation \cite{masubuchi2020deep,nguyen2024two,uslu2025maskterial,uslu2024open,nguyen2025varphiadaptphysicsinformedadaptationlearning,pandey2025cliff,zhu2020deformable,carion2020end,redmon2016you,ronneberger2015u,kirillov2023segment,ravi2024sam}. Related work on robust visual perception has explored continual learning, adaptive learning, and domain generalization under changing classes, domains, and acquisition conditions. These directions include continual semantic scene understanding and multimodal continual learning \cite{truong2023fairness,truong2025falcon,truong2026phidpo}, continual and fairness-aware domain adaptation \cite{truong2024conda,truong2023fredom}, and broader adaptive and domain-generalized perception \cite{truong2024eagle,nguyen2022selfsupervised,truong2022otadapt,truong2021bimal,truong2020unvp}.

More recently, physics-aware multimodal large language models (MLLMs), such as Physics-Aware Instruction Tuning (QuPAINT) \cite{nguyen2026qupaintphysicsawareinstructiontuning}, combine visual localization with reasoning grounded in the optical properties of the material and substrate. These models provide richer outputs than conventional detectors, including localized candidate regions, material-layer predictions, and natural-language explanations.

\noindent \textbf{Observation to Action.}
Localization is only one part of the characterization workflow. After candidate flakes are identified, researchers may ask how many were returned, which satisfy a criterion, which is largest, what their physical dimensions are, or where selected candidates lie in the original image. These operations require reliable manipulation of the expert's returned observations.

This becomes difficult when the analysis is expressed as long free-form text containing many spatial coordinates rather than as an explicit scientific state. Summary quantities can become inconsistent with the underlying enumeration, and placing a general-purpose language model in front of the expert does not resolve the problem because the orchestrator must still count, filter, rank, or measure observations in context.

We study this behavior using QuPAINT as a fixed domain expert. Across 228 fully reviewed images, its candidate enumeration varies systematically with image density, while summary counts from longer outputs can become inconsistent with that enumeration. In-Context OpenQlaw exhibits the same limitation, and controlled replay shows that structured representation alone does not reliably resolve it. Exact aggregation is obtained when the operation itself is delegated to a deterministic function.

Based on these findings, we present \textbf{OpenQlaw}, a tool-using agentic AI assistant for 2D quantum material analysis. A general-purpose vision-language orchestrator delegates material characterization to QuPAINT and exact downstream operations to dedicated tools. In State-Grounded OpenQlaw, each expert response is converted into an image-scoped observation record bound to the source image and expert reply. Counting, selection, ranking, calibrated bounding-box measurement, and rendering are performed over the identified record, with explicit refusal when required information is unavailable.

\noindent \textbf{Related Work in Tool-Augmented Agents.}
Delegating computation to external tools is well established. Program-aided language models use interpreters for exact arithmetic \cite{gao2023pal}, while tool-augmented agents integrate external operations into their reasoning processes \cite{yao2023react,schick2023toolformer}. Scientific agents such as EAA \cite{du2026eaa} apply vision--language models to automate microscopy workflows, emphasizing instrument control rather than deterministic computation over expert-generated observations.
OpenQlaw builds on these foundations but addresses a distinct problem: reliably computing over observations already produced by a domain expert. State-Grounded OpenQlaw binds expert responses to their source images through hashed, image-scoped records with explicit observation scopes. This enables verification of the source image and response, the scope used for computation, and refusal conditions involving cross-image requests or missing calibration. It also separates aggregation errors from errors in the underlying candidate list, allowing each to be evaluated independently. Our structured replay baseline isolates the benefit of structured observations without introducing state-managed tool execution. We note that OpenQlaw preserves QuPAINT's original analysis while making downstream operations explicit and deterministic. This allows disagreement between the returned candidate list and the reviewed image-level reference to be studied separately from errors introduced during aggregation.

\noindent \textbf{Contributions.} There are three main contributions of this work. First, we identify and characterize a downstream aggregation limitation that arises when long, free-form scientific outputs are used directly for exact operations, and show that the same behavior appears when a general-purpose orchestrator operates over the expert response. Next, we introduce State-Grounded OpenQlaw, using image-scoped observation records and deterministic scientific operations that preserve image, reply, and scope identity while supporting counting, selection, ranking, calibrated bounding-box measurement, rendering, and explicit refusal conditions. Lastly, through controlled replay and live evaluation, we show that structured representation alone does not reliably resolve aggregation, while deterministic delegation removes counting error over the selected record and makes candidate-list/reference and scope discrepancies independently observable.
\section{The OpenQlaw Framework}
\label{sec:framework}

\begin{figure}[t!]
    \centering
    \includegraphics[width=1\textwidth]{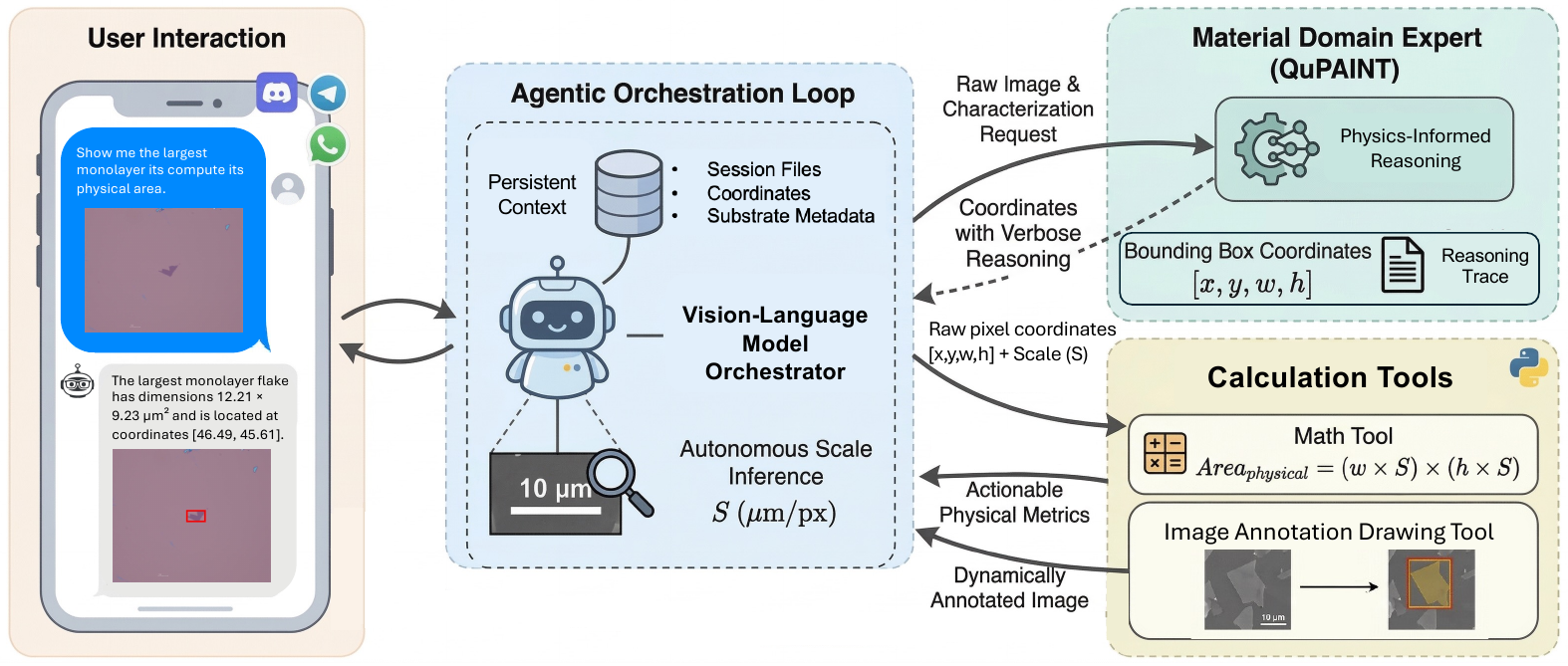}
    \caption{Overview of OpenQlaw. The orchestrator delegates material characterization to the domain expert, converts the returned observations into an image-scoped record, and applies deterministic scientific operations over the identified record.}
    \label{fig:OpenQlaw_framework}
\end{figure}

OpenQlaw is a tool-using agentic AI system that connects a general-purpose vision-language orchestrator with a physics-aware domain expert and deterministic scientific operations. The orchestrator interprets the researcher's request, delegates material characterization to the expert, and applies scientific tools over the returned observations. This preserves the expert's original analysis while separating it from downstream operations that require exact manipulation of its output. \Cref{fig:OpenQlaw_framework} provides an overview.

\noindent \textbf{Agentic Orchestration.}
OpenQlaw supports interaction through messaging interfaces such as WhatsApp and Discord, allowing researchers to provide optical microscopy images together with natural-language requests. During a session, the orchestrator can retain the active image, returned observations, and experimental metadata for follow-up queries such as selecting the largest candidate or requesting a calibrated measurement.

Requests requiring material-specific image analysis are delegated to the domain expert. Requests requiring operations over previously returned observations can instead invoke deterministic scientific tools. The orchestrator therefore interprets user intent and selects the appropriate component, while domain analysis and exact downstream computation remain separated.

\begin{figure}[t]
    \centering
    \includegraphics[width=.85\textwidth]{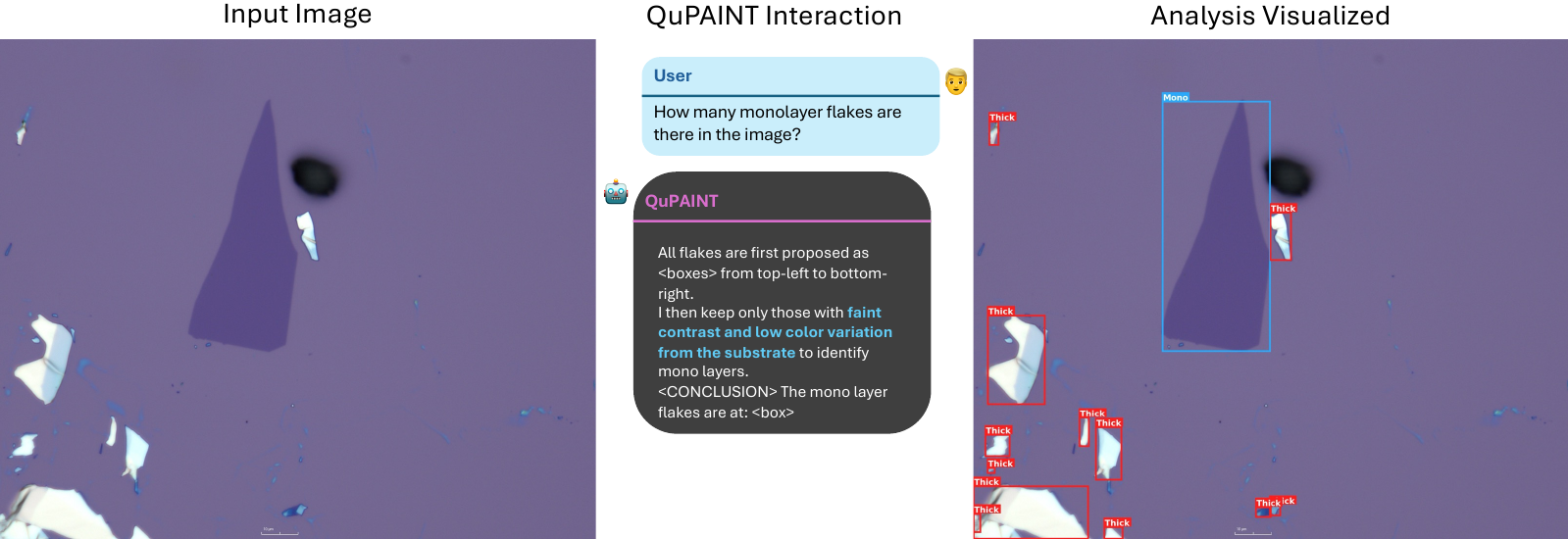}
    \caption{Example of QuPAINT flake analysis. From an optical microscopy image and user query, QuPAINT localizes candidate flakes and identifies monolayer candidates using optical contrast and substrate-relative color cues.}
    \label{fig:qupaint_example}
\end{figure}

\noindent \textbf{Material Domain Expert.}
We use QuPAINT \cite{nguyen2026qupaintphysicsawareinstructiontuning} as the Material Domain Expert. Given an optical microscopy image and characterization request, QuPAINT performs physics-aware material analysis and returns localized candidate flakes as bounding boxes in whole-image percentage coordinates $[x,y,w,h]$, together with natural-language reasoning and a final conclusion. \Cref{fig:qupaint_example} illustrates a representative interaction and the corresponding localized candidate analysis.

QuPAINT is kept fixed throughout this work using a single pinned checkpoint. OpenQlaw does not modify its model parameters or inference procedure, allowing us to study downstream operations over the returned observations while preserving the expert's original analysis.

\subsection{In-Context Operations}
\label{sec:baseline}

With In-Context OpenQlaw, the orchestrator receives the expert response as text and is instructed to use the returned analysis for operations such as counting and filtering. These operations are therefore performed directly by the language model over the expert's free-form output. This approach preserves the full expert response, but exact operations must still be inferred from a potentially long enumeration of observations. In-Context OpenQlaw provides no deterministic operation over the returned candidate set, motivating the State-Grounded formulation below.

\subsection{State-Grounded Operations}
\label{sec:records}
State-Grounded OpenQlaw converts each expert response into an explicit observation record without altering the original analysis. The host parses and validates the returned coordinates, then exposes deterministic operations over the resulting record.

\noindent\textbf{Image-Scoped Observation Records.}
Each valid bounding box is stored in an \texttt{AnalysisRecord} bound to the source image and expert response using SHA-256 hashes. The \emph{proposals} scope contains all valid enumerated boxes, while the \emph{conclusion} scope contains boxes within a well-formed final \texttt{<CONCLUSION>} block. Malformed, out-of-image, and exact duplicate boxes are excluded, while distinct overlapping boxes are preserved rather than merged without additional evidence.

\noindent\textbf{Deterministic Scientific Operations.}
The \texttt{openqlaw\_scientific\_record} tool supports five operations over an identified record. \emph{Count} returns the number of observations in a scope, \emph{Select} filters using available layer labels, \emph{Rank} orders observations by bounding-box area, \emph{Measure} computes physical bounding-box areas in $\mu$m$^2$ when image-specific calibration is available, and \emph{Render} draws explicitly selected object identifiers on the verified source image. QuPAINT outputs bounding boxes in image-relative percentage coordinates, which the host converts to pixel coordinates using the source image dimensions before applying any physical calibration. Bounding-box area is not interpreted as segmented flake area.

\noindent\textbf{Explicit Refusals.}
The tool refuses requests involving an ungranted record identity, unavailable layer labels, missing image calibration, unspecified render targets, or identities from different images. Calibration remains bound to the image for which it was provided.

\noindent\textbf{Tool-Directed Output.}
The expert response is returned without modification. The host appends a typed footer identifying the available records and directing the orchestrator to the scientific tool when deterministic computation is required. Record size is not exposed, so exact counts must be obtained through the corresponding operation.
\section{Implementation}
\label{sec:implementation}

OpenQlaw is implemented using NanoBot \cite{hkuds2026nanobot}, an open-source agent framework inspired by OpenClaw \cite{steinberger2026openclaw}. The orchestrator uses Qwen3-VL-32B-Instruct \cite{yang2025qwen3,bai2025qwen3vltechnicalreport} at a fixed model revision and is served with vLLM using \texttt{bfloat16} precision and a $32K$ context window. QuPAINT is served independently through FastAPI and exposed through the Model Context Protocol (MCP). The orchestrator and domain expert execute on separate GPUs and communicate locally through the tool interface. The deployed system supports WhatsApp and Discord. For evaluation, the messaging layer is bypassed and the orchestrator is driven directly using frozen inputs to ensure consistent experimental conditions.

\noindent\textbf{Tool Integration.}
QuPAINT is registered as the \texttt{get\_flake\_data} tool and provides the material-analysis output used for downstream interaction. The \texttt{draw\_specific\_flakes} tool supports rendering selected observations. In State-Grounded OpenQlaw, \texttt{openqlaw\_scientific\_record} provides deterministic access to the image-scoped records described in \Cref{sec:records}. After a QuPAINT call, the host parses the returned coordinates, constructs the corresponding \texttt{AnalysisRecord}, and appends the typed record information before returning the expert response to the orchestrator. The original QuPAINT output is preserved without modification.

\noindent\textbf{Evaluation Harness.}
We record provider requests and responses, tool calls, and tool results for each trial. Experimental campaigns use frozen prompts and image manifests with SHA-256 hashes, together with write-once traces and environment records containing the model configuration, expert checkpoint, and hardware used for execution. One recorded retry is permitted for transport failures. We do not perform semantic retries or regenerate outputs based on their content. Trials that violate the experimental protocol remain in the trace and are excluded from the corresponding analysis rather than removed. 
\section{Experimental Setup}
\label{sec:setup}

\noindent \textbf{Data and Evaluation References.}
We evaluate OpenQlaw using WSe$_2$ and WS$_2$ optical microscopy images from a materials laboratory database. Images have a resolution of $3,248\times2,900$ pixels and were acquired at 25.2$\times$ magnification with a recorded scale of 0.1087\,$\mu$m/px.
Each crystallite entry contains a bounding box, material classification, and human review status. Because individual reviewed entries do not imply exhaustive image-level review, we define an image as \emph{fully reviewed} only when every associated crystallite entry has been reviewed. These images provide an exhaustive reference for the reviewed material-flake count. However, the reference does not establish a one-to-one correspondence between each returned QuPAINT proposal and a physical flake, so we do not interpret count disagreement as proposal-level detection error.
Completeness is therefore defined by the laboratory's per-image review protocol and does not guarantee identification of flakes absent from the annotations.

The database contains 7{,}251 fully reviewed images, including 2{,}870 with at least one material flake. Of 14{,}287 reviewed material flakes, 14{,}218 are labeled thick, and only 69 are mono-, bi-, or trilayer. We therefore evaluate count agreement at scale, while layer-based selection and calibrated measurement are evaluated through tool behavior and refusal conditions.

We use a 20-image WSe$_2$ pilot for live evaluation of In-Context and State-Grounded OpenQlaw. The pilot supports reference-free measurements including aggregation behavior, tool use, scope selection, and refusal handling. We evaluate count agreement with the reviewed image-level reference only on the fully reviewed cohort described below.

We construct a cohort of 228 fully reviewed images containing 160 WSe$_2$ and 68 WS$_2$ images, excluding the pilot. Images are selected deterministically using seed 20260913 and stratified by reviewed material-flake count to cover sparse through densely populated images.

\noindent \textbf{Evaluation Conditions.}
We evaluate three system configurations and one controlled replay setting.
The \emph{direct expert} condition queries QuPAINT without the OpenQlaw orchestrator and characterizes the relationship between its returned enumeration and summary count.

\emph{In-Context OpenQlaw} represents the original workflow described in \Cref{sec:framework}. The orchestrator receives the QuPAINT response as free-form text and performs downstream operations directly in context.

\emph{State-Grounded OpenQlaw} uses the image-scoped records and deterministic operations described in \Cref{sec:records}. We evaluate two live variants. Variant A retains the same orchestrator policy used by In-Context OpenQlaw, while Variant B additionally includes an explicit instruction to use the scientific tool.

The \emph{matched replay} provides a controlled comparison between representation and delegation using identical archived QuPAINT observations. The orchestrator receives either the raw expert response, the same validated observations as a structured record, or the structured record with access to a deterministic count function. Record identifiers are opaque, and record size is withheld, preventing the target count from being exposed directly.
Replay uses the same orchestrator model as OpenQlaw but a minimal interaction loop rather than the complete NanoBot workflow. This allows us to compare representation and aggregation while holding the underlying expert observations fixed.

\noindent \textbf{Metrics and Evaluation Protocol.}
For each trial, we record exact agreement with the requested record count, agreement with the expert enumeration, tool use, record identity and scope, and refusal handling. We additionally compare candidate-list size with the reviewed image-level count on the fully reviewed cohort.
Images are the primary evaluation unit, with repeated comparisons summarized by the per-image median. Paired comparisons in matched replay use 10{,}000 seeded bootstrap samples over images.

Prompts and image manifests are frozen before execution, and semantic retries based on response content are not permitted. Live OpenQlaw runs use two NVIDIA L40S GPUs for the orchestrator and one L40S for QuPAINT, while replay uses four NVIDIA A100-40GB GPUs. Because the live and replay configurations use different hardware, they are evaluated separately rather than treated as bit-identical executions.

All experiments use a fixed user request (``Count all distinct material flakes in this micrograph. End with exactly one line: COUNT=$<$nonnegative integer$>$.''), greedy decoding, and predetermined random seeds (20260913 for cohort selection and 20260912 for replay, with three paired repeats). All 2{,}052 replay trials and 684 expert trials were completed without exclusions. The 20-image pilot was excluded from the evaluation cohort by design.
\section{Results}
\label{sec:results}
We first evaluate QuPAINT directly on the 228 fully reviewed images. Candidate-list size strongly tracks the reviewed image-level flake count ($\rho=0.91$), but the summary count stated in the free-form response agrees with its own enumeration in only 13\% of responses where a count is present. Producing candidate observations and aggregating them into an exact summary are therefore distinct operations.

In-Context OpenQlaw exhibits the same limitation. On the 20-image pilot, the orchestrator matches the expert enumeration in only 6 of 60 trials and the conclusion count in 2 of 60, while 52 match neither. The median absolute difference from the expert enumeration is 6 observations, and 43\% of returned counts are multiples of five. In one trial, for example, QuPAINT returns 97 candidate observations and the orchestrator reports 100.

\subsection{In-Context Counting Degradation}
We isolate representation from deterministic delegation using matched replay. Identical QuPAINT observations are presented as raw text, as a validated structured record, or with access to the deterministic count operation used by State-Grounded OpenQlaw. Each condition contains 684 trials from 228 images across three repeated runs.

Raw and structured replay return the exact record count in 40\% and 42\% of trials, respectively, while State-Grounded OpenQlaw is exact in 684/684 trials. The paired improvement over structured replay is $+0.58$ with a 95\% bootstrap interval of $[0.52,0.64]$. By comparison, structured representation improves over raw replay by only $+0.02$ with an interval of $[-0.03,0.08]$, showing that representation alone does not reliably improve exact aggregation.

\begin{figure}[t!]
    \centering
    \includegraphics[width=0.75\textwidth]{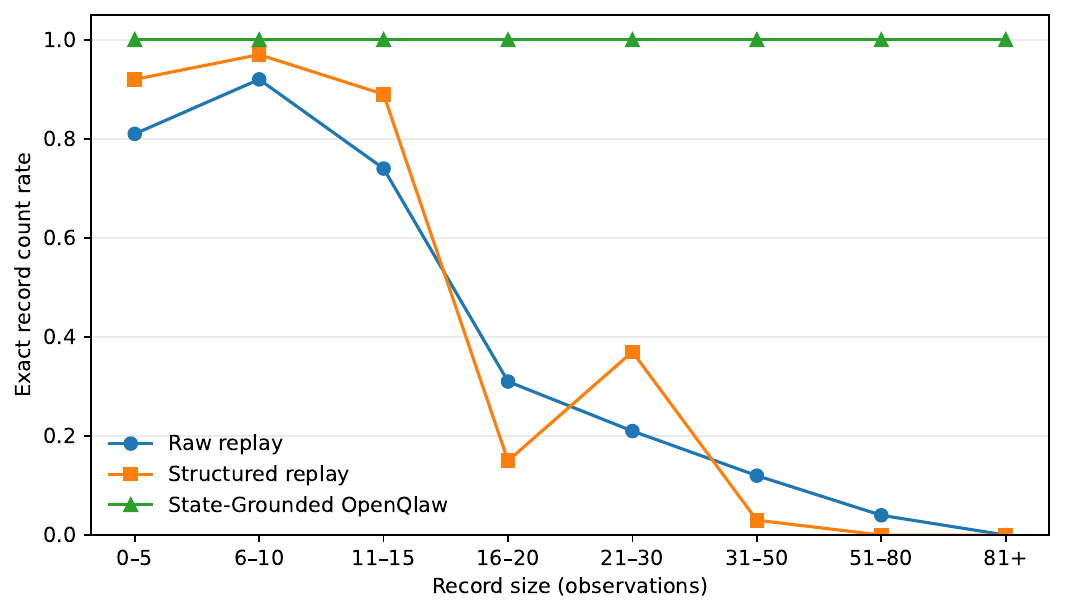}
    \caption{Exact record-count rate by record size in matched replay. In-context counting degrades sharply beyond approximately 15 observations, while State-Grounded OpenQlaw remains exact across all record-size groups.}
    \label{fig:record_size}
\end{figure}

As shown in \Cref{fig:record_size}, exact in-context counting declines sharply beyond approximately 15 observations. At 16--20 observations, exact agreement falls to 31\% for raw replay and 15\% for structured replay. At 51--80 observations, the rates fall to 4\% and 0\%.
Longer records increasingly produce estimates rather than exact enumeration. For 31--50 observations, median returned-to-record ratios are 1.04 for raw replay and 0.92 for structured replay, falling to 0.95 and 0.85 for 51--80 observations. The same behavior appears for WSe$_2$ and WS$_2$.

\subsection{State-Grounded OpenQlaw}
We next evaluate State-Grounded OpenQlaw within the complete live workflow. \Cref{tab:live_tools} summarizes tool adoption, count consistency, scope selection, and refusal behavior. Variant A retains the In-Context OpenQlaw policy without an additional tool-use instruction, yet invokes the scientific tool in 60/60 trials, with every reported count matching the tool result. Variant B adds an explicit tool-use instruction but invokes the tool in 58 of 60 trials. Both versions are State-Grounded OpenQlaw, but we compare the effect of an explicit tool-use instruction to the agent policy.


\begin{table}[t]
\centering
\caption{Live State-Grounded evaluation on the 20-image pilot (60 trials per variant). Since the request specifies all flakes, \emph{proposals} is the intended scope. Reported counts matched tool outputs in all tool-using trials.}
\label{tab:live_tools}
\scriptsize
\setlength{\tabcolsep}{5pt}
\begin{tabular}{lcccc}
\toprule
Variant & Tool Use $\uparrow$ & Scope (Prop./Concl.) & Intended Scope $\uparrow$ & Refusals $\uparrow$ \\
\midrule
A & \textbf{60/60} & 48 / 12 & 48/60          & \textbf{2/2} \\
B & 58/60          & 49 / 9  & \textbf{49/60} & \textbf{1/1} \\
\bottomrule
\end{tabular}
\end{table}

The explicit record further distinguishes aggregation errors from scope selection errors. Since the locked request asks for all flakes, the intended scope is \emph{proposals}, selected in 48 of 60 trials for Variant A and 49 of 60 for Variant B. The remaining tool-using trials (12 and 9, respectively) instead select the \emph{conclusion} scope, which QuPAINT restricts to candidate monolayers. Variant B also produces two responses without invoking the tool. Across all tool-using trials, the reported count matches the selected record exactly, indicating that the remaining errors arise from scope selection rather than aggregation. In In-Context OpenQlaw, these failure modes are indistinguishable from the final count alone. Explicit records instead associate each answer with its source record and scope, allowing scope selection errors to be identified directly from the execution trace.\\

\subsection{Isolating Candidate-List Discrepancy}
Let $\hat{n}$ denote the returned count, $r$ the selected record size, and $n$ the reviewed image-level flake count. Their difference decomposes as
\[
\hat{n}-n=(\hat{n}-r)+(r-n),
\]
where $\hat{n}-r$ is the aggregation discrepancy and $r-n$ is the candidate-list/reference discrepancy. State-Grounded OpenQlaw reduces the first term to zero in all 684 matched replay trials, allowing the remaining discrepancy to be examined independently.

Candidate-list size remains strongly correlated with the reviewed image-level count ($\rho=0.91$), although the relationship varies with image density. These differences should not be interpreted as proposal-level detection errors because the reviewed references do not establish a one-to-one correspondence between returned boxes and physical flakes. Distinct overlapping proposals are therefore preserved rather than merged without additional evidence.

For the six cohort images in which selected-record size equals the reviewed image-level count, State-Grounded OpenQlaw also returns the reviewed count in 6/6 cases. Candidate enumeration itself remains variable across repeated QuPAINT calls. This variability remains upstream, while downstream computation over any identified record remains exact.

\subsection{Interactive Use}
\begin{figure}[t!]
    \centering
    \includegraphics[width=.85\textwidth]{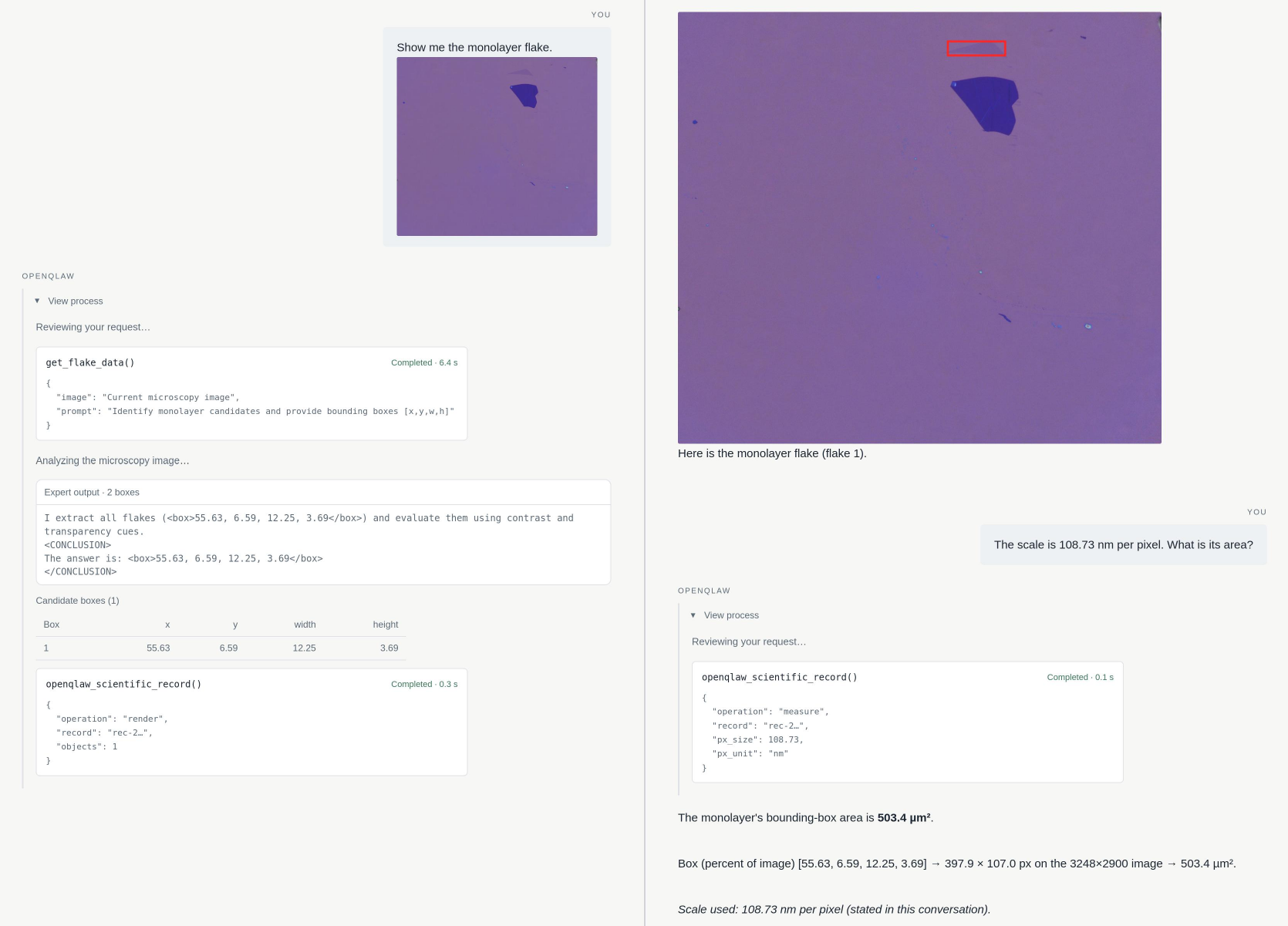}
    \caption{\textbf{State-Grounded OpenQlaw Interaction.} QuPAINT returns bounding boxes, which are converted to pixels using the source image dimensions. \emph{Measure} then applies the user-provided spatial calibration to compute bounding-box area in $\mu$m$^2$.}
    \label{fig:OpenQlaw_interaction}
\end{figure}

While the controlled experiments bypass the user interface, the deployed system supports interactive analysis through a web interface, WhatsApp, and Discord. \Cref{fig:OpenQlaw_interaction} illustrates an interaction in which QuPAINT identifies a candidate flake, the observation is rendered from its associated record, and a subsequent request computes its calibrated area from the same record.

\section{Discussion}
\label{sec:discussion}

These results demonstrate that reliable scientific interaction depends not merely on structuring model-generated observations, but on delegating subsequent operations to deterministic computation.
This separation also clarifies where remaining disagreement originates. Once aggregation is exact over an identified record, disagreement with the reviewed image-level count can no longer be attributed to an additional counting error. It instead reflects the returned candidate list or the selected scope. Because the reviewed annotations do not establish a one-to-one correspondence between every returned proposal and a physical flake, we do not interpret candidate-list disagreement as proposal-level detection error. Multiple distinct observations may correspond to the same physical region, and OpenQlaw preserves overlapping proposals rather than merging them without additional evidence.

Explicit records similarly make scope selection observable. The orchestrator may correctly operate over the narrower \emph{conclusion} scope when the intended target is the full \emph{proposals} scope. By binding downstream operations to an identified record and scope, State-Grounded OpenQlaw separates aggregation correctness from scope selection and candidate-list disagreement. This makes each source of discrepancy independently measurable rather than entangled within a single free-form response.

\noindent\textbf{Limitations.}
This work evaluates one general-purpose orchestrator, one QuPAINT checkpoint, and optical microscopy images from one laboratory setting at a single magnification. The reviewed annotations are strongly dominated by thick flakes, so counting is evaluated more extensively than layer-specific selection. Calibrated measurement, ranking, selection, and rendering are verified through tool behavior and refusal conditions rather than large-scale live accuracy studies.

Matched replay uses a minimal interaction loop and different hardware from the live OpenQlaw system. It therefore supports controlled comparison of representation and delegation, while the live experiments evaluate tool adoption, scope selection, and refusal behavior. Measurements in this work refer to bounding-box area rather than segmented flake area, and the reviewed references do not provide proposal-level ground truth.
\section{Conclusion}
In this work, we have studied the reliability of downstream operations over model-generated scientific observations. We have shown that structured representation alone does not ensure consistent results, while deterministic delegation has achieved exact record counts across all trials. We have introduced State-Grounded OpenQlaw, which organizes a domain expert's (QuPAINT) outputs into image-scoped records and performs downstream operations through deterministic computation. This approach preserves QuPAINT's scientific analysis while separating aggregation errors from scope selection and candidate-list discrepancies, providing a reliable interface between multimodal reasoning and exact scientific operations.

\noindent\textbf{Limitations.}
Our evaluation is limited to a single general-purpose orchestrator, a fixed QuPAINT checkpoint, and optical microscopy images acquired under one laboratory setting and magnification. The dataset is heavily dominated by thick flakes, restricting the assessment of layer-specific operations. Furthermore, selection, ranking, calibrated measurement, and rendering are evaluated through deterministic tests and live refusal checks rather than extensive accuracy experiments.
Candidate lists may also differ across repeated expert calls. While deterministic delegation does not resolve this variability, it makes its source identifiable by grounding each answer in a specific record. Differences can thus be attributed to the underlying observations rather than aggregation errors. 

\noindent\textbf{Future Work.}
Future work will study generalization across orchestrators, materials, and imaging conditions while preserving explicit scientific state and deterministic downstream computation.



%
%
%
\begin{credits}
\subsubsection{\ackname}
This work is partly supported by MonArk NSF Quantum Foundry (DMR-1906383), NSF PCL-Test Bed AURORA (2607535) and  NSF Quantum Award (2444042). It acknowledges the Arkansas High-Performance Computing Center for providing GPUs.

\subsubsection{\discintname}
The authors have no competing interests to declare that are relevant to the content of this article.
\end{credits}
\bibliographystyle{splncs04}
\bibliography{references}
\end{document}